# Proposing a Localized Relevance Vector Machine for Pattern Classification

Farhood Rismanchian[1,*], Karim Rahimian[2]

[1]Department of Information and Industrial Engineering, Yonsei University, Seoul, South Korea
[2]Department of Computer sciences, University of Economic Sciences, Tehran, Iran

**Abstract**—Relevance vector machine (RVM) can be seen as a probabilistic version of support vector machines which is able to produce sparse solutions by linearly weighting a small number of basis functions instead using all of them. Regardless of a few merits of RVM such as giving probabilistic predictions and relax of parameter tuning, it has poor prediction for test instances that are far away from the relevance vectors. As a solution, we propose a new combination of RVM and *k*-nearest neighbor (*k*-NN) rule which resolves this issue with regionally dealing with every test instance.

In our settings, we obtain the relevance vectors for each test instance in the local area given by *k*-NN rule. In this way, relevance vectors are closer and more relevant to the test instance which results in a more accurate model. This can be seen as a piece-wise learner which locally classifies test instances. The model is hence called localized relevance vector machine (LRVM). The LRVM is examined on several datasets of the University of California, Irvine (UCI) repository. Results supported by statistical tests indicate that the performance of LRVM is competitive as compared with a few state-of-the-art classifiers.

1. Introduction

Linear regression models are appealing for their simplicity, high interpretability, and cheap computational cost [1, 4, 6]. Assume a training data set with $N$ input pair vectors $\{\mathbf{x}_i, t_i\}_{i=1}^N$ where $\mathbf{x}$ is input vector and $t$ is target value. The goal is to find predictions upon function like $y(\mathbf{x})$ over the input space via training weights $\mathbf{w}$.

For more accurate predictions and stronger fit, the data is mapped to the feature space which may help learner to make data linearly separable in higher dimensions. However, this approach suffers from overfitting problem. Because many generated features may be irrelevant and hence it is good to prune irrelevant features from learning process [2, 3]. The process of pruning is called sparsity in the machine learning community.

There are several researches conducted sparsity techniques such as least absolute shrinkage and selection operator (LASSO) solution [35] as a sparse regression treatment to the overfitting problem wherein the L-1 norm of output weights is added to the objective function least square errors which is L-2 norm. Another popular sparsity approach is to have a statistical view to the linear regression problem and hence a Bayesian treatment to overfitting problem. Indeed, the function $y(\mathbf{x})$ can be estimated in feature space as follows:

$$y(\mathbf{x}) = \sum_{i=1}^{M} w_i \phi_i(\mathbf{x}) = \mathbf{w}^T \phi(\mathbf{x}) \tag{1}$$

where $\phi_i$ is the $i$-th basis function among $M = N + 1$ basis functions, drawn from a Gaussian distribution and $\mathbf{w}$ is unknown weight vector. As a Bayesian treatment for linear regression model, one can directly formulate Eq.1 with assuming priors on the weights $\mathbf{w}$.

As an example of sparse Bayesian learning (SBL) we call relevance vector machine (RVM) [4] which produces highly sparse solutions by pruning irrelevant training features with assigning priors on $\mathbf{w}$. Finding sparse solutions is performed by placing Gaussian priors with mean zero and variance $\mathbf{A}$ over weights $\mathbf{w}$. RVM is relax of tuning sparsity parameter. Indeed, the rate of sparseness is automatically determined during the training process.

Although the percentage of sparsity is significant for RVM and this helps it to be very quick during run time, i.e. the time for prediction of unseen data, this also may be problematic. The issue of significant sparseness rate in RVM is, some test instances may be far away from the basis functions, and hence RVM has poor predictions for those test instances [11, 12]. This is because of the covariance matrix in RVM which is degenerate. Degeneracy means the covariance matrix is low rank [12, 13].

In this study, instead globally sparsifying the data, we locally do it. We only use $k$ number of instances surrounds the test instance. We find the instances using $k$-nearest neighbor ($k$-NN) rule. This is performed for every test instance. The proposed model works only on local areas obtained by $k$-NN. This can be seen as a piecewise variant of RVM, is hence called localized RVM (LRVM).

The rest of this study is organized as follows. Related works, motivations and contributions are discussed in next section. Section 3 is for preliminaries. Section 4 describes the proposed method and section 5 is related to experimental results conducted on several datasets. Section 6 discusses the main points of the proposed LRVM algorithm. Finally, last section concludes remarks.

2. Related works

There are a few improvements on the accuracy and speed of RVM that we discuss in this section.

In [12], researchers try to get a constant prior variance as a replacement for zero values for diagonal of the covariance matrix which is generated by the covariance function. This is performed by the covariance function normalization. This normalization changes the shape of basis functions. However, the normalization brings long-term correlations which are away from basis centers. It means, despite making non-zero (constant) the diagonal of the covariance matrix, it is still low rank. To resolve this problem, priors are decorrelated by adding a white noise Gaussian process to the model right before normalization. This setting could enrich the prior over functions and hence could improve RVM.

Rasmussen et.al [13] addresses the issue of degeneracy of covariance function in RVM and then heal RVM by an additional basis function centered at the test input. This new basis function is introduced only at test time and is used for every test input. The manipulation of the covariance function using an additional basis function brings computational burden to the classic RVM.

In [30], a new regression method for continuous estimation of the intensity of facial behavior interpretation is formulated, which is called Doubly Sparse Relevance Vector Machine (DSRVM). DSRVM enforces double sparsity by jointly selecting the most relevant training examples or relevance vectors and the most important kernels.

In [31], a novel methodology for spatial prediction of landslides on the basis of the relevance vector machine classifier (RVMC) and the cuckoo search optimization (CSO) is invesitgated. The RVMC is used to generalize the classification boundary that separates the input vectors of landslide conditioning factors into two classes: landslide and non-landslide.the new approach employs the CSO to fine-tune the basis function's width used in the RVMC. A geographic information system (GIS) database has been established to construct the prediction model.

Study at [32] introduces a novel sparse Bayesian machine-learning algorithm for embedded feature selection in classification tasks. The model is called the relevance sample feature machine (RSFM) and is able to simultaneously choose the relevance instances and also the relevance features for regression or classification problems. Experimental comparisons on synthetic as well as benchmark data sets show that RSFM is successful in both feature selection and classification.

In [33], one of RVM drawbacks which is the lack of an explicit prior structure over the weight variances which can lead to severe overfitting or oversmoothing is addressed. Hence, an efficient scheme to control sparsity in Bayesian regression by incorporating a flexible noise-dependent smoothness prior into the RVM is introduced. An empirical evaluation of the effects of choice of prior structure and the link between Bayesian wavelet shrinkage and RVM regression are presented. According to their result, they could outperform RVM performance in terms of goodness of fit and achieved sparsity as well as computational performance.

In [34], a novel kernel is introduced which is called adaptive spherical Gaussian kernel. It is used for nonlinear regression, and the stagewise optimization algorithm for maximizing Bayesian evidence in RVM. Extensive experiments on artificial datasets and real-world datasets shows its effectiveness and flexibility on representing regression problem with higher levels of sparsity and better performance than classical RVM. The interesting point of the model is, to automatically choose the right kernel widths locally fitting RVs from the training dataset, which could keep right level smoothing at each scale of signal.

### 2.1 Motivations and Contributions

With above review in place, we are interested in improving RVM working on the problem of RVM mentioned at [12, 13]. To do so, we are inspired by a paper in the literature which improves support vector machine (SVM) and is called localized SVM (LSVM) [16]. LSVM builds a linear SVM model for each test instance using only the training instances which surround that test instance. The goal is using a linear SVM than nonlinear one to not deal with kernel parameters of nonlinear SVM. Our proposed LRVM follows similar approach like LSVM. In our study, we address the problem of the degeneracy of covariance function in RVM [12, 13] using $k$-NN rule. Another research as a source of inspiration in this study is, the combination of RVM and polynomial learners, called sparse Bayesian reduced Polynomial (SBRP) proposed at [36]. The way of sparsity using RVM could help to enrich the efficiency of Polynomial learner. Bringing sparsity into Polynomial model could make the proposed SBRP insensitive to the Polynomial order. Thus, we are also interested to verify if our proposed LRVM is insensitive to $k$, i.e. the number of neighbors.

The contributions of this paper are as follows:
1- LRVM deals with only on a small portion of train data (for every test instance) instead of whole data. Estimating inverse of a small covariance matrix during learning process is more accurate than computing inverse of a large covariance matrix coming from the whole data [15]. So, this may be a reason for superiority of LRVM to RVM in terms of final classification accuracy. Another merit of LRVM over RVM is, every test instance in LRVM has its own relevance vectors (RVs). This means we no longer have the issue if a test instance is far away from RVs. This helps to improve the final classification accuracy.

2- LRVM needs to create a kernel for neighbors of every test instance and this is timely. To speed up this process, we make a look-up table. We only make once the kernel for whole training data. For every test instance, the index of neighbors from each row and column of this kernel matrix is considered instead making a local kernel for neighbors of that test instance.

3- Extensive experiments, following a statistical test, conducted on several datasets of UCI data repository validate the effectiveness of the LRVM compared to a few states of the arts in terms of classification accuracy.

3. Preliminaries

3.1. *k*-NN

The *k*-NN became popular since 1967 and it is still one of the most favorable pattern classification methods thus far [5, 14, 18] and among ten top data mining algorithms [14]. *k*-NN is an instance-based learner wherein the target value of every query instance is approximated regionally using a majority vote of its neighbors. The target value of a query data assigned to the class most common among its *k* nearest neighbors [5,18]. The complexity of k-NN is O(*LN*) for *N* training data points with *L* dimensions. As *k* increases, the complexity increases as well.

3.2. RVM

Relevance vector machine (RVM) [4, 9, 10, 29] is a sparse Bayesian learner (SBL) which can be seen as a probabilistic variant of SVM with fewer basis functions. Prediction in RVM follows Eq. 1, i.e. $\mathbf{y} = \mathbf{w}^{\mathrm{T}}\mathbf{\Phi}$, where $\mathbf{y}$ is a predictor vector, $\mathbf{w}$ is a weight vector and $\mathbf{\Phi}$ is a set of basis functions of training instances. For classification task, every training instance $\mathbf{x}$ is assumed that is drawn from an independent Bernoulli random variable with probability $p(\mathbf{t} \mid \mathbf{x})$ where $\mathbf{t}$ is a target vector. The posterior probability $p(\mathbf{w}, \boldsymbol{\alpha} \mid \mathbf{t})$ of class membership of $\mathbf{x}$ is objective function which should be maximized. The Bernoulli likelihood for training instances is given as follows:

$$p(\mathbf{t} \mid \mathbf{w}) = \prod_{i=1}^{N} \sigma\{g(\mathbf{k}_i; \mathbf{w})\}^{t_i}[1 - \sigma\{g(\mathbf{k}_i; \mathbf{w})\}]^{1-t_i} \qquad (2)$$

where $\mathbf{k}_i$ is *i*-th column vector of the kernel matrix $\mathbf{K}$ and g is the network output and $\sigma$ is the Sigmoid function to force outputs to be in the range of [0, 1]. Due to the nonlinearity of

Sigmoid function, a closed form solution is impossible and an iterative procedure, called Mackay should be performed.

To obtain the weight parameter, the posterior probability $p(\mathbf{w}, \boldsymbol{\alpha} \mid \mathbf{t})$ is decomposed to obtain the marginal likelihood $p(\mathbf{w} \mid \mathbf{t}, \boldsymbol{\alpha})$ which is the key component of optimization in RVM. Again, the marginal likelihood $p(\mathbf{w} \mid \mathbf{t}, \boldsymbol{\alpha})$ cannot be directly solved and hence should be decomposed as:

$$p(\mathbf{w} \mid \mathbf{t}, \boldsymbol{\alpha}) \propto p(\mathbf{t} \mid \mathbf{w}) p(\mathbf{w} \mid \boldsymbol{\alpha}) \tag{3}$$

To impose sparsity, one can assume $p(w_i \mid \alpha_i) \approx N(0, \alpha_i^{-1})$ for every instance $\mathbf{x}_i$. Furthermore, marginalization technique is used to taking integral to integrate out $\mathbf{w}$ in Eq. (3), however the generated integral is interactable and Laplace approximation is performed, and finally an Iterative Regularized Least Square (IRLS) is used to solve the posterior mode $\mathbf{w}$, denoted as $\hat{\mathbf{w}}$. Two unknowns $\hat{\mathbf{w}}$ and $\boldsymbol{\Sigma}$ (coming from the term $p(\mathbf{t} \mid \mathbf{w})$) can be solved as follows:

$$\hat{\mathbf{w}} = \boldsymbol{\Sigma} \mathbf{K}^{\mathrm{T}} \mathbf{B} \mathbf{t} \tag{4}$$

$$\boldsymbol{\Sigma} = (\mathbf{K}^{\mathrm{T}} \mathbf{B} \mathbf{K} + \mathbf{A})^{-1} \tag{5}$$

where $\mathbf{A} = diag(\boldsymbol{\alpha})$ and $\mathbf{B}$ is a diagonal matrix with elements $\beta = \sigma\{y(x_n)\}[1 - \sigma\{y(x_n)\}]$ (details can be found at [4, 9]). Since this is an iterative procedure, given $\hat{\mathbf{w}}$ and $\boldsymbol{\Sigma}$, the hyperparameters $\alpha_i$ can be updated as,

$$\alpha_i^{new} = \frac{1 - \alpha_i \Sigma_{ii}}{\hat{w}_i^2} \tag{6}$$

where $\hat{w}_i$ is the $i$-th posterior weight from Eq. (4), and $\Sigma_{ii}$ is the $i$-th diagonal element of $\boldsymbol{\Sigma}$ from Eq. (5). The parameters $\hat{\mathbf{w}}$, $\boldsymbol{\Sigma}$, and $\alpha_i$ from Eqs. (4), (5), and (6) are optimized in every iteration until we reach a stopping condition. The key point in RVM iterative procedure is, in early iterations many $\alpha$'s tend to infinity which means the corresponding w's tend to zero and hence the corresponding features are pruned from basis set and are excluded for next iteration. A test instance $\mathbf{x}_i^{test}$ can be predicted as $y_i^{test} = \mathbf{w}^{\mathrm{T}} \boldsymbol{\Phi}(\mathbf{x}_i^{test})$ wherein $\mathbf{w} \neq 0$ is only used.

## 4. Proposed method
### 4.1. Degeneracy of RVM and *k*-NN solution

As it was mentioned, finding sparse solutions in RVM is performed by placing Gaussian priors with mean zero and variance $\mathbf{A}$ over weights $\mathbf{w}$. Indeed, there is an inverse relation between weights $\mathbf{w}$ and its variance $\mathbf{A}$. As $\mathbf{A}$ goes to infinity, $\mathbf{w}$ tends to zero. The goal is controlling weights **w** by learning variances of the weights, **A** [4]. With this setting, RVM has a drawback. Taking the Gaussian priors over the weights with *N* number of training instances and *M* = *N*+*1* number of basis functions ('1' is for bias term) with covariance matrix $\Sigma = \Phi_{NM} A \Phi_{NM}^T$ makes the covariance matrix degenerate (Note that $\Phi_{NM}$ is design matrix.). Degeneracy means the covariance matrix is low rank. The maximum rank for this matrix is the number of bases after sparsification or the size of relevant features which is very small. The core of the problem stems from the covariance function $k(\mathbf{x}_i, \mathbf{x}_j)$ which is degenerate [12]:

$$k(\mathbf{x}_i, \mathbf{x}_j) = \emptyset^T(\mathbf{x}_i) \mathbf{A} \emptyset(\mathbf{x}_j) \tag{7}$$

where $\mathbf{x}_i$ and $\mathbf{x}_j$ are input vectors and $\emptyset(\mathbf{x})$ is Gaussian basis/kernel function for input **x**. We should see how the covariance function generates the prior variance which is placed in the diagonal of covariance matrix $\Sigma$ where $\mathbf{x}_i = \mathbf{x}_j$:

$$k(\mathbf{x}, \mathbf{x}) = \sum_{m=1}^{M} \alpha_m \emptyset_m^2(\mathbf{x}) \tag{8}$$

If we move away from the basis functions, its mean value tends to zero and the prior variance $k(\mathbf{x}, \mathbf{x})$ becomes smaller and smaller. In another words, the model uncertainty is high around the centers of relevance vectors (RVs) and is very low as we move away from the centers or RVs while we expected to get larger uncertainty (by increasing variance) as we further move away from the centers. Thus, one can say RVM is overconfident about predictions for those test instances are far away from the basis functions [11, 12].

In this study, we regionally address this issue using *k*-NN method. We only use *k* number of instances or neighbors for every test instance. With new settings $M \leq k$ where *k* is the number of training instances which are similar to test data. We use the term "similarity", because it is the covariance function that defines nearness or similarity [11]. We use only few instances in design matrix $\Phi$ and this reduced design matrix is used to make the covariance matrix $\Sigma$. So, the proposed model less suffers from the degeneracy problem.

### 4.3. LRVM: Nearest neighbor approach to RVM

Let $\{(\mathbf{x}_1, y_1), ..., (\mathbf{x}_N, y_N)\}$ be train set, where $\mathbf{x}_i$ is the *i*-th train instance and $y_i$ is its corresponding target values. For every test instance, *k* neighbors are chosen from train set using a similarity metric like Euclidean distance. For every test instance $\mathbf{x}_i^{test}$, there is a sphere with volume $\mathbf{V}_i$ which surrounds *k* points in train set those are nearest neighbors as we have shown in

Fig. 1.

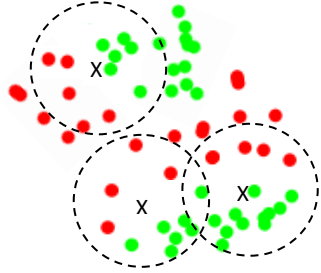

**Fig. 1**. Apply LRVM for every test instance

The dashed circles are volumes $\mathbf{V}_i$ which surround $k$ neighbors for every test point $\mathbf{x}_i^{test}$ which is shown by cross lines in Fig. 1. The $k$ points of $\mathbf{V}_i$ are $k$-rows of the train matrix in input space. In next step of LRVM algorithm, $\mathbf{V}_i$ is mapped to kernel space as follows,

$$Neighb(\mathbf{x}_i^{test}) = \begin{bmatrix} \emptyset_{1,1} & \cdots & \emptyset_{k,1} \\ \vdots & \ddots & \vdots \\ \emptyset_{1,k} & \cdots & \emptyset_{k,k} \end{bmatrix} \quad (9)$$

The purpose is regional sparsification by RVM approach. Simply, for each localized space $\mathbf{V}_i$, we find sparse solutions. The general diagram of LRVM is shown in Fig. 2.

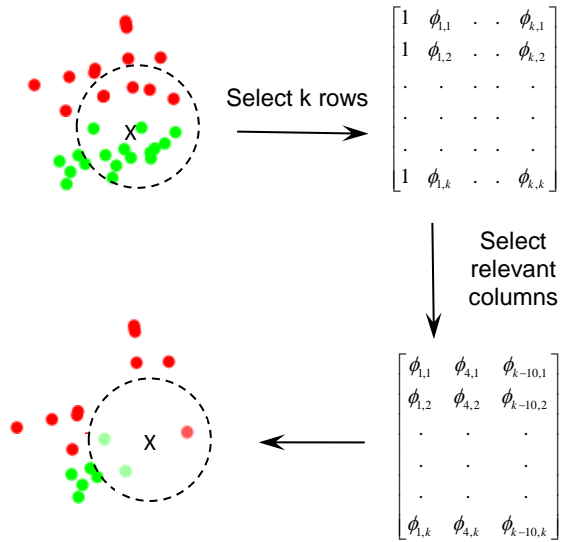

**Fig. 2**. Application of the proposed LRVM on a test instance

Referring to the Fig. 2, first, the $k$ rows of the training data for that test instance in input space is mapped to feature space. Next, RVM method is performed to obtain the most relevant instances and their corresponding weights in the mapped space. These weights are between zero and one and that is why we showed them with low resolution in bottom left of Fig . 2. These weights are used to determine the class of the test instance $\mathbf{x}_i^{test}$ (the crossed-sign in the dashed circle of Fig. 2) through $y_i^{test} = \mathbf{w}^T \mathbf{\Phi}(\mathbf{x}_i^{test})$ as explained in last paragraph in section 3.2. This algorithm is performed for every test instance $\mathbf{x}_i^{test}$.

An important point that comes to mind is if we need to create kernel submatrix for every set of neighbors for every test instance? This is timely. A simple solution is to only for one time create a Gram matrix for all training data and store the index of every training instance of this Gram matrix. For every test instance, once we find its *k* nearest neighbors, we fetch the index of the neighbors and only corresponding *k* rows and *k* columns of this neighbors are called and selected from the large Gram matrix than re-computation of kernel for that *k* neighbors. This way we can save time for the proposed LRVM.

## 5. Results

### 5.1. Data and setup

For empirical evaluation of the proposed LRVM model, it is implemented on 20 datasets of UCI data repository. Table 1 shows their descriptions.

Table 1. Descriptions of UCI datasets*

| no | Collection name | No. of instances | No. of attributes (R/I/N)* | No. of Classes |
|---|---|---|---|---|
| 1 | Wbcd | 699 | 9 (0/9/0) | 2 |
| 2 | Australia | 690 | 14(8/0/6) | 2 |
| 3 | Titanic | 2201 | 3 (0/3/0) | 2 |
| 4 | Heart | 270 | 13 (1/12/0) | 2 |
| 5 | Bands | 365 | 19 (13/6/0) | 2 |
| 6 | Ionosphere | 351 | 33 (32/1/0) | 2 |
| 7 | Pima | 768 | 8 (8/0/0) | 2 |
| 8 | Bupa | 345 | 6 (1/5/0) | 2 |
| 9 | Shuttle | 279 | 6 (0/9/0) | 2 |
| 10 | Parkinson | 197 | 23 (23/0/0) | 2 |
| 11 | Sonar | 208 | 60 (60/0/0) | 2 |
| 12 | Iris | 150 | 4 (4/0/0) | 3 |
| 13 | Wine | 178 | 13 (13/0/0) | 3 |
| 14 | Balance | 625 | 4 (4/0/0) | 3 |
| 15 | Nursery | 12960 | 8 (8/0/0) | 4 |
| 16 | Zoo | 101 | 16 (0/0/16) | 7 |
| 17 | Segment | 2310 | 19 (19/0/0) | 7 |
| 18 | Ecoli | 336 | 7 (7/0/0) | 8 |
| 19 | Pendigit | 10992 | 16 (0/16/0) | 10 |
| 20 | Optdigit | 5620 | 64 (0/64/0) | 10 |

*R: real, I: integer, N: nominal*

Some notes for the experiments' setup are as follows:
1. The feature normalization is applied before finding *k* nearest neighbors. Each feature is normalized as follows:

$$\hat{f}_j = \frac{f_j - \mu_j}{\sigma_j} \qquad (10)$$

   Where $f_j$ is the *j*-th feature vector, and $\mu_j$ and $\sigma_j$ are the mean and standard deviation of the feature *j* in the training set.

2. For the sake of comparison, the performance of the proposed LRVM is compared with a few classifiers such as the support vector machines with Gaussian kernel (SVM-GK) chosen from the LIBSVM proposed in [19], linear SVM in the form of SVMrank [20], a relevance vector machine with Gaussian kernel (RVM-Gauss) [4], a relevance vector machine with a "Bernoulli" kernel (RVM-Bern) [4], and a sparse variant of extreme learning machine (ELM) [21] called sparse pseudoinverse-ELM or SPI-ELM [26], and a reduced multivariate polynomial model with total error rate as objective function (TER-RM), proposed at [22], *k*-NN and fuzzy-NN (F-NN).

3. The sigmoid function for hidden layer of SPI-ELM is $G(a,b,x) = \frac{1}{1+e^{-(ax+b)}}$, where ***a***, ***b*** are the random synapses and bias with uniform distribution within $[-1,1]$. Additionally, to evaluate the performance of SPI-ELM, the number of hidden neurons H should be tuned. The range for H in our experiment is $[10, 20, \ldots, 250]$.

4. For SVM, the hyperparameters $[c, \gamma]$ with regulator *c* and Gaussian kernel width $\gamma$ is conducted on points $\{2^{-3}, 2^{-2}, \ldots, 2^{6}\} \times \{2^{-3}, 2^{-2}, \ldots, 2^{6}\}$ and for RVM with Gaussian kernel, the width parameter is also in the same range.

5. For *k*-NN, the value for *k*, the number of nearest neighbors is in the range $[1, 3, \ldots, 71]$. For LRVM, even values are also added to this range. To report the model accuracy, we use the best value of *k* to predict the test classes. To find the best value, 10-fold cross validation is used and the mean value of accuracies for all folds is taken into consideration.

6. For TER-RM, the range for choosing regularization term *b* is $\{10^{-4}, 10^{-3}, \ldots, 10^{2}\}$, and the order for polynomial kernel is $\{1, 2, 3, 4\}$.

7. The experiments are conducted in MATLAB environment running on a 3.40-GHz-CPU with 32-GB RAM.

Table 2 shows the average classification accuracy (CA) on twenty UCI datasets. We report the average CA for nine classifiers using 10 runs of 10-fold stratified cross validation.

**Table 2**. Average classification accuracy for Nine Classifiers over twenty datasets

|   | Model / Data set | $k$-NN | F-NN | SVM-rank | SVM-GK | SPI-ELM | RVM-G | RVM-B | TER-RM | LRVM |
|---|---|---|---|---|---|---|---|---|---|---|
| 1 | WBCD | 0.931 | 0.942 | 0.971 | 0.974 | 0.968 | 0.681 | 0.952 | 0.975 | **0.987** |
| 2 | Australia | 0.859 | 0.864 | 0.853 | 0.870 | 0.867 | 0.840 | 0.831 | 0.8705 | **0.893** |
| 3 | Titanic | 0.751 | 0.756 | 0.723 | 0.772 | 0.755 | 0.775 | 0.768 | 0.7750 | **0.815** |
| 4 | Heart | 0.826 | 0.822 | 0.821 | 0.843 | 0.857 | 0.848 | 0.805 | 0.8428 | **0.87** |
| 5 | Bands | 0.7115 | 0.718 | 0.703 | 0.733 | 0.741 | 0.728 | 0.719 | 0.7414 | **0.789** |
| 6 | Iono | 0.873 | 0.883 | 0.875 | **0.944** | 0.894 | 0.856 | 0.841 | 0.9116 | 0.941 |
| 7 | Pima | 0.687 | 0.688 | 0.734 | 0.763 | **0.82** | 0.743 | 0.787 | 0.7744 | 0.809 |
| 8 | Bupa | 0.669 | 0.667 | 0.692 | 0.711 | 0.70 | 0.714 | 0.710 | 0.735 | **0.816** |
| 9 | Shuttle | 0.995 | 0.992 | 0.973 | 0.981 | 0.982 | 0.985 | 0.988 | 0.978 | 0.986 |
| 10 | Parkinson | 0.945 | 0.945 | 0.893 | 0.952 | 0.875 | 0.934 | 0.921 | **0.956** | 0.951 |
| 11 | Sonar | 0.830 | 0.830 | 0.853 | **0.872** | 0.84 | 0.619 | 0.785 | 0.787 | 0.862 |
| 12 | Iris | 0.982 | 0.982 | 0.975 | **0.987** | 0.97 | 0.936 | 0.949 | 0.972 | 0.982 |
| 13 | Wine | 0.920 | 0.932 | 0.958 | 0.972 | 0.89 | 0.964 | 0.964 | **0.992** | 0.990 |
| 14 | Balance | 0.964 | 0.959 | 0.945 | 0.977 | 0.946 | 0.953 | 0.928 | 0.934 | **0.984** |
| 15 | Nursery | 0.873 | 0.88 | 0.893 | 0.925 | 0.937 | 0.905 | 0.917 | 0.915 | **0.980** |
| 16 | Zoo | 0.885 | 0.90 | 0.888 | 0.927 | 0.91 | 0.914 | 0.943 | 0.987 | **0.989** |
| 17 | Segment | 0.878 | 0.883 | 0.924 | 0.934 | 0.954 | 0.873 | 0.886 | 0.949 | **0.987** |
| 18 | Ecoli | 0.883 | 0.904 | 0.888 | **0.959** | 0.914 | 0.896 | 0.888 | 0.887 | 0.955 |
| 19 | Pendigit | 0.972 | 0.981 | 0.925 | 0.985 | **0.995** | 0.945 | 0.958 | 0.974 | 0.984 |
| 20 | Optdigit | 0.967 | 0.930 | 0.926 | 0.974 | **0.989** | 0.944 | 0.943 | 0.968 | 0.985 |
|  | Average | 0.870 | 0.873 | 0.871 | 0.903 | 0.89 | 0.853 | 0.874 | 0.896 | **0.929** |
|  | Average rank | 6.55 | 6.8 | 6.65 | 2.8 | 4.5 | 6.45 | 6.6 | 2.9 | **1.75** |

*Bold values indicate the best value under identical conditions.*

Referring to results in Table 2, LRVM is competitive to SVM-GK, TER-RM and SPI-ELM and is better than other chosen classifiers according to the CA as the chosen performance measure.

In our experiments, the best value of $k$ for LRVM is often 1 or 2 or 3 for all datasets which means that the LRVM is insensitive to the change of number of neighbors while $k$ is a sensitive parameter for both $k$-NN and f-NN. This is in line with the study at [36], a combination of RVM and Polynomial network which is called SBRP. They showed that SBRP is insensitive to the Polynomial order.

To verify which model is statistically significant, a non-parametric Friedman test including Nemenyi diagram [23] which is used for comparison of multiple classifiers over multiple datasets is adopted. The purpose is to statistically compare all classifiers over all UCI datasets.

The Friedman test is a rank-based algorithm which ranks the classifiers for each dataset separately. The best performing algorithm gets the rank of 1, the second best gets the rank of 2 and so forth. Let assume $r_i^j$ is the rank of $j^{th}$ of G algorithms on the $i^{th}$ of $L$ datasets. The goal is to compare the average rank of algorithms $R_j$. The null hypothesis tests whether there is significant difference among algorithms or not.

If the fisher value $F_f$ is larger than the critical value $CV_\alpha$ obtained from fisher-distribution (can be found in any statistical book) with $(G-1)$ and $(G-1)(L-1)$ degrees of freedom and with $\alpha = 0.05$, the null hypothesis is rejected. The equations to compute Fisher value $F_f$ are as follows [23]:

$$\chi_F^2 = \frac{12L}{G(G+1)} \left[ \sum_j R_j^2 - \frac{G(G+1)^2}{4} \right] \quad (11)$$

Then:

$$F_f = \frac{(L-1)\chi_F^2}{L(G-1) - \chi_F^2} \quad (12)$$

The null-hypothesis is rejected if $F_f > CV_\alpha$. After rejection, we can proceed with a Nemenyi post-hoc test which is useful to compare classifiers. The performance of two classifiers is significantly different if the corresponding average ranks for them differ by at least the critical difference (CD). It means if the difference between two algorithms is greater than CD, the algorithms are significantly different CD can be found via:

$$CD = CV_\alpha \sqrt{\frac{G(G+1)}{6L}} \quad (13)$$

In our experiments, the critical value for the two-tailed Nemenyi test $CV_{0.05}$ is 3.102 and the Fisher value is 12.2717. Since $F_f > CV_\alpha$ is satisfied, the null hypothesis is rejected.

The CD value for AUC is 2.6864 and the average ranks for AUC are as the last row of Table 2. To better observe and compare classifiers, Nemenyi diagram is shown in Fig. 3.

Referring to the Nemenyi post hoc test in Fig. 3 which is just applied after Friedman test, there are three different sets of classifiers (the thick horizontal bars). Those classifiers are not significantly different are connected. According to the Nemenyi test, the performance of two classifiers is significantly different if their ranks are higher than at least the critical difference (CD). Moreover, the two classifiers have no significant difference when they are in the same group. The obtained CD value is 2.6864. LRVM, SVM-GK, and TER-RM are in same group and are not significantly different. Other classifiers are in two other groups. It can be seen that LRVM is located in the top rank group and the difference of average rank values between SPI-ELM and LRVM is higher than CD value. Hence, LRVM is statistically different from SPI-ELM.

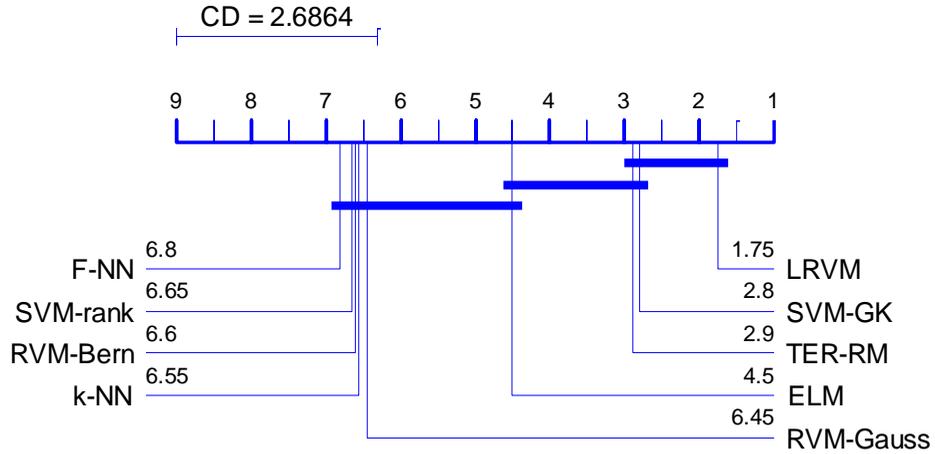

Fig. 3. Comparing classifiers via the Nemenyi diagram.

## 6. Discussion

To visually compare RVM and LRVM, we apply them on Ripley's synthetic data in only two dimensions which is a binary classification problem. Fig. 5 shows the decision boundary for RVM and local neighborhood for LRVM, and also the relevance vectors (RVs), and local relevance vectors (LRVs). The width of Gaussian kernel for both RVM and LRVM is set to 0.5.

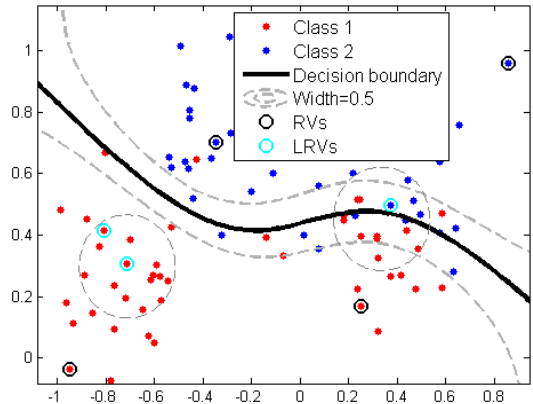

Fig. 4. RVM and LRVM classifications on Ripley's synthetic data.

Referring to Fig. 4, every test instance in LRVM has its own local relevance vectors (LRVs) chosen from training data. The LRVs (centers of basis functions) are very close to the test instances. If we wanted to classify these instances by RVM, we had to use information of too far RVs.

traditional RVM requires a few iterations to update hyperparameters and find sparse solutions. However, in LRVM, we deal with a small number of instances (small submatrix) for every test point. So, we found using very few iterations for this small submatrix can also give us the best sparse solutions which maximize the marginal likelihood.

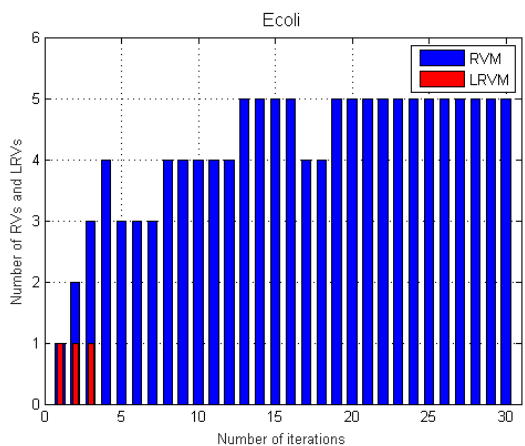

**Fig. 5**. The number of iterations for RVM and LRVM versus the number of RVs and LRVs.

Fig. 5 shows the bar plot of the number of iterations for both RVM and LRVM models. As it can be seen the LRVM model only requires three iterations and one LRV to be terminated (for a randomly chosen test point) while we need thirty iterations and five RVs to terminate RVM. We considered termination criteria to be when no update for hyperparameter $\alpha$ seems worthwhile [10]. Indeed, the sequential learning for RVM proposed at [10] begins with only one basis function and then does three actions: add, delete or update hyperparameters. So, in LRVM for this randomly selected test instance, after three iterations, model cannot do any action and for the case of RVM, model stops after thirty iterations.

## 7. Conclusion

Improving the performance of RVM is the main purpose of this study. This was performed by solving the issue of RVM regarding poor predictions for those test instances that are far away from the relevance vectors. We regionally solved this problem working on a small size of training data obtained by *k*-NN for every test instance. Instead making sparse the entire training data, we sparsify the subspace of *k* neighbors of every test instance. Proposed LRVM automatically finds irrelevant dimensions and assigns their weights to zero with an assumed prior distribution. This brings sparsity and high generalization capability to the model.

We could speedup training of LRVM using look-up table along with decreased number of iterations. Our future research direction can be embedding RVM and *k*-NN to improve the performance of k-NN. This way can bring the least squares estimations in RVM to the prototype selection of *k*-NN.